\documentclass[letterpaper]{article} % DO NOT CHANGE THIS
\usepackage[submission]{aaai23}  % DO NOT CHANGE THIS
\usepackage{times}  % DO NOT CHANGE THIS
\usepackage{helvet}  % DO NOT CHANGE THIS
\usepackage{courier}  % DO NOT CHANGE THIS
\usepackage[hyphens]{url}  % DO NOT CHANGE THIS
\usepackage{graphicx} % DO NOT CHANGE THIS
\usepackage{natbib}  % DO NOT CHANGE THIS AND DO NOT ADD ANY OPTIONS TO IT
\usepackage{caption} % DO NOT CHANGE THIS AND DO NOT ADD ANY OPTIONS TO IT
\usepackage{algorithm}
\usepackage{algorithmic}

\usepackage{newfloat}
\usepackage{listings}
\DeclareCaptionStyle{ruled}{labelfont=normalfont,labelsep=colon,strut=off} % DO NOT CHANGE THIS
\floatstyle{ruled}
\newfloat{listing}{tb}{lst}{}
\floatname{listing}{Listing}
\title{Time Series Prediction for Food Sustainability}
\author{
    Fiona Victoria Stanley Jothiraj\textsuperscript{\rm 1}
}
\affiliations{
    \textsuperscript{\rm 1}University of Washington Bothell, USA\\
    
    fiona123@uw.edu
}

\usepackage{bibentry}
\begin{document}

\maketitle

\begin{abstract}
With exponential growth in the human population, it is vital to conserve natural resources without compromising on producing enough food to feed everyone. By doing so, we can improve people's livelihoods, health, and ecosystems for the present and future generations. Sustainable development, a paradigm of the United Nations, is rooted in food, crop, livestock, forest, population, and even emission of gases. By understanding the overall usage of natural resources in different countries in the past, it is possible to forecast the demand in each country. The proposed solution consists of implementing a machine learning system using a statistical regression model that can predict the top-k products that would endure a shortage in each country in a specific period in the future. The prediction performance in terms of absolute error, root mean square error show promising results due to its low errors. This solution could help organizations and manufacturers understand the productivity and sustainability needed to satisfy the global demand.
\end{abstract}

\section{Introduction}
According to the first principle of the 1992 Rio Declaration on Environment and Development (United Nations 1992), "Humans are at the center of concern for sustainable development. They are entitled to a healthy and productive life that is in harmony with nature." With over 7.9 billion humans, it is getting harder for the majority of the population to lead a healthy life. Around 9.9\% of the population, which accounts for 811 million people, still go to bed on an empty stomach. On the contrary, over 1.3 billion tonnes of food are wasted every year. The world's population is rapidly growing, and it is estimated that there will be around 10 billion people on Earth by the year 2050.
Environmentalists have been trying to find solutions to reduce the numbers in terms of hunger and food wastage. Sustainable food development ensures that the current and future human population has enough food to eat and access high-quality, nutritious foods.
Transitioning to a sustainable food system involves understanding the demand for a particular crop/food product required in each country. The EAT-Lancet Commission (Willett et al. 2019) proposed five strategies as general starting points for national, regional, city, and local change as part of sustainable food development,
    \begin{itemize}
         \item Seek international and national commitment to shift toward healthy diets
         \item Reorient agricultural priorities from producing high quantities of food to producing
        healthy food
         \item Sustainably intensify food production to increase high-quality output.
         \item Strong and coordinated governance of land and oceans
         \item At least halve food losses and waste, in line with UN Sustainable Development Goals
    \end{itemize}
With these strategies in mind, a means of knowing the demand for a particular crop/food product required in each country would be beneficial to humans.
The problem can be addressed by developing a machine learning system that could predict the food products that would endure shortages in a particular area at a specific time. The solution uses a vector autoregressive (VAR) statistical model that captures the influence of the production of crops and livestock due to the emission of greenhouse gases. The result of the system is quite simple to comprehend since it lists out the top-k products and its production trend graphs. The limitation of the proposed approach is that the solution only accounts for a single environmental factor - the emission of gases to forecast the production of crops, livestock, or forestry products.

\section{Literature review}
Extensive research has been performed in the field of machine learning for social science to discover new findings, understand the causal effects, and make predictions. Scholars have experimented with various traditional mathematical models, machine learning models and deep learning methods for food demand forecasting. Some of the popular choices include ARIMA, Holt-Winters, supervised
regression models, and artificial neural networks like NARXNN (non-linear auto regressive exogenous neural network).

The research (Lutoslawski et al. 2021) uses a nonlinear autoregressive neural network for food demand prediction. The solution consists of data collected from over 3.5 years trained with NARXNN. The optimal models provided forecasts with the lowest mean absolute error (MAE), mean absolute percentage error (MAPE), and root mean squared error (RMSE) values. 

The proposed prediction model (Ren et al. 2007) is used to predict the region yield for wheat-based on crop biomass estimation. It uses a net primary production model to estimate crop biomass. 

The paper on price prediction for agricultural products (Chen et al. 2019) uses a long short-term memory (LSTM) model using wavelet analysis for prediction. The experiments show that this model achieved better performance and accuracy. 

The papers mentioned above use neural networks to solve food demand prediction. However, not much research has been done using statistical machine learning models like VAR. The proposed solution aims to solve this challenge by using the vector autoregressive regression model to predict the production of crops and livestock.

\section{Data Preparation}
    \subsection{Dataset}
    The dataset for the solution is provided by Food and Agriculture Organization Corporate Statistical Database (FAOSTAT). The database provides free access to food and agriculture data for over 245 countries and territories and covers all FAO regional groupings from 1961 to the most recent year available (2019). The three datasets chosen from this database are as follows,
        \begin{itemize}
            \item Climate Change: Emissions Total
            \item Forestry: Forestry Production and Trade
            \item Production: Crops and livestock products
        \end{itemize}

    \subsubsection{1. Climate Change: Emissions Total}
    The dataset summarizes the emissions generated from agricultural and forest land. The gas emissions consist of methane (CH4), nitrous oxide (N2O), and carbon dioxide (CO2) emissions
     
    from sources like crops, livestock, forest management, and land use. Emissions data in the units of kilo-tonnes are available for each country for the period 1961-2019. The dataset consists of over a million instances.
    
    \begin{center}
        \includegraphics[width=0.45\textwidth]{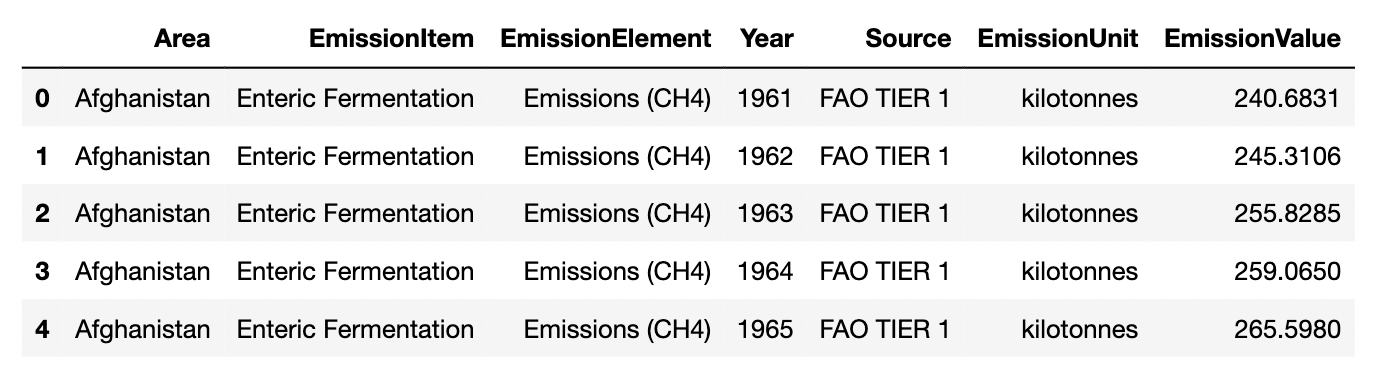}
        \captionof{figure}{Sample of Emissions dataset}
    \end{center}

    \subsubsection{2. Forestry: Forestry Production and Trade}
    The dataset contains information on the production and trade in primary wood and paper products for all countries and territories worldwide. The primary forest products included in this database are Roundwood, sawnwood, wood-based panels, pulp, and paper and paperboard. The dataset consists of over two million instances.

    \begin{center}
        \includegraphics[width=0.45\textwidth]{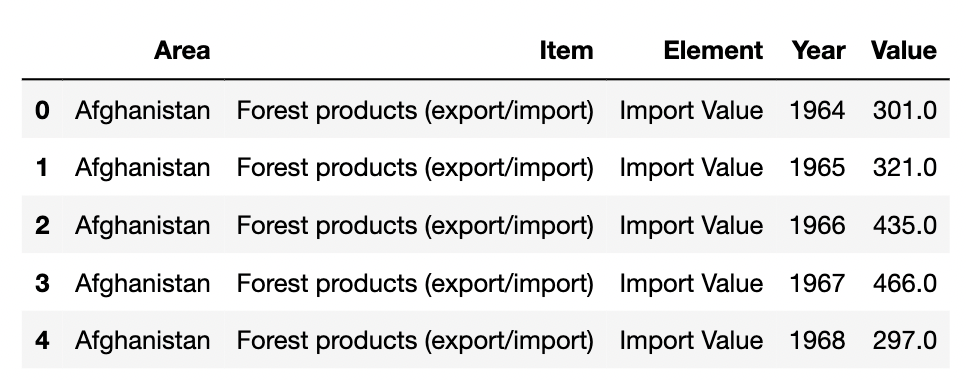}
        \captionof{figure}{Sample of Forestry production dataset}
    \end{center}

    \subsubsection{3. Production: Crops and livestock products}
    The dataset contains crop and livestock production statistics for all countries and regions for over 173 items. The items include Crops Primary, Fibre Crops Primary, Cereals, Coarse Grain, Citrus Fruit, Fruit, Jute Jute-like Fibres, Oilcakes Equivalent, Oil crops Primary, Pulses, Roots and Tubers, Tree Nuts and Vegetables and Melons. The data is expressed in area harvested, production quantity, and yield. There are over 3.8 million instances found in the database.
    
    \begin{center}
        \includegraphics[width=0.45\textwidth]{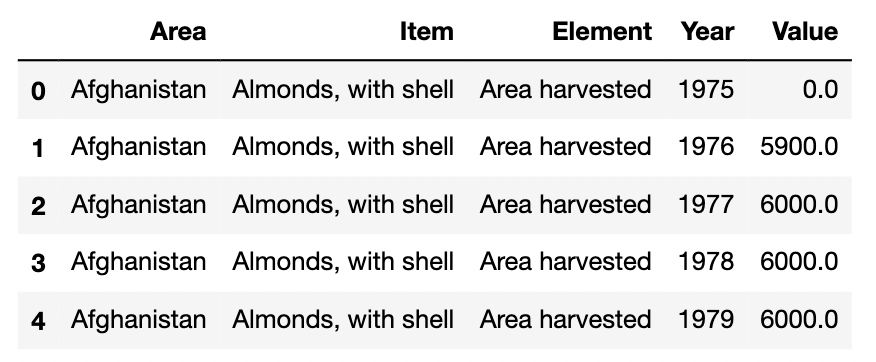}
        \captionof{figure}{Crops and livestock products dataset}
    \end{center}

    \subsection{Preprocessing}
    Since the data size is too large, the dataset was explored more to deal with extensive data. It is evident that each country's data is independent of the other, i.e., the production of grapes in India would not affect the production of apples in China. This makes it possible to train each country's data separately, significantly reducing the data size from 300 million to ~1 million. Each country's data is individually trained with its emission, production, and forestry datasets.
    
    The steps involved in preprocessing of data are as follows:
    \begin{enumerate} 
        \item Removing unwanted features like Area, Source and Emission Unit
        \begin{description}
            \item a. Area - The feature is removed since the country that is being used for prediction is already known
            \item b. Source - This feature is removed since the values in these rows like ‘FAO TIER 1’ provides no valuable information that is required for forecasting
            \item c. Emission Unit - The feature is removed since the emission unit in terms of kilotonnes is common throughout the dataset
        \end{description}
        \item Removing redundant instances \\
    Case 1:
    Table 1 shows three sample instances from the production crop dataset. Although there are three instances for the production of carrots, namely, yield, area harvested, and production, each element can be calculated from the other.
    \\ Production = 800 tonnes = 800 * 10,000 = 8,000,000 hg
    \\ Production / Area harvested = 8,000,000 / 120 = 66667 hg/ha = Yield 

    \begin{table}[htb]
    \centering
    %\resizebox{.95\columnwidth}{!}
    
    \begin{tabular}{l|l|l|l|l}
        Country & Year & Crop & Element & Value \\
        Panama & 1961 & Carrots & Yield & 66667 hg/ha\\
        Panama & 1961 & Carrots & Area harvested & 120 ha\\
        Panama & 1961 & Carrots & Production & 800 tonnes\\
    \end{tabular}
    \caption{Production crop sample data.}
    \end{table} 
    
    Case 2:
    Similarly in Table 2, Indirect emission + Direction emission = Emission (CH4) 0.005100 + 0.015700 = 0.020800. This helps us understand that the indirect and direct instances (or) total emission could be removed since they add no value to the model. The presence of redundant instances increases the data size. Using domain knowledge, the number of useful instances present in the data sets can be improved.

    \begin{table}[htb]
        \centering
        %\resizebox{.95\columnwidth}{!}
        \begin{tabular}{l|l|l|l}
            Country & Year & Emission & Value (kilo tonnes) \\
            Panama & 1961 & Indirect (CH4) & 0.0051\\
            Panama & 1961 & Direct (CH4) & 0.0157\\
            Panama & 1961 & Total emission (CH4) & 0.0208\\
        \end{tabular}
        \caption{CH4 Emission sample data.}
    \end{table} 
    
    \item Removing features having NaN values \\
    In this step, features consisting of at least one NaN value are completely scraped from the dataset. This decision was made after performing the following two steps,\\
    \textit{Experiment 1:}\\
    Initially, experimentation was done by only removing features that are about 30\% of NaN values. The mean imputation imputed the missing values. However, this method was wrong since it did not account for the factor of being a time series, and the imputed values did not make much sense.\\
    \textit{Experiment 2:}\\
    Using forward or backward fills, the linear interpolation method was used to impute the missing values. This method did not fit the problem statement since multiple features had data from 1961-2000, and several features had missing values for 2000 - 2019. The inconsistency in the dataset meant that one type of linear interpolation would not work for all circumstances.\\
    
    \item Pivot Table \\
    A pivot table is used to reorganize the data stored in a table. Data that essentially consists of features like ‘Emission Element’, ‘Emission Amount’, ‘Production crop’,’ Year’ etc. was converted to tables with respect to the Year and Emission/Production crop. The modified dataset is shown in Figure 4 and Figure 5. Figure 4 is a pivot table of a concatenated dataset consisting of a production crop and forestry databases.
    
    \begin{center}
        \includegraphics[width=0.45\textwidth]{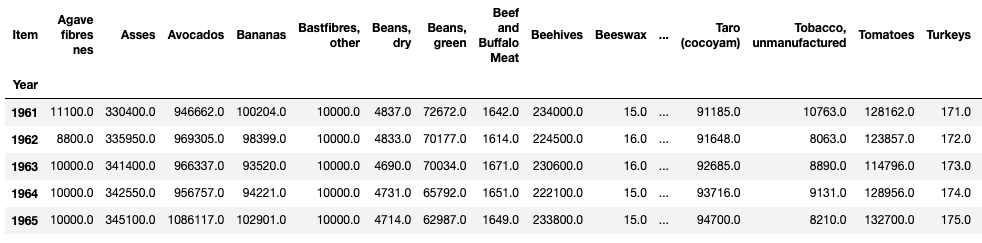}
        \captionof{figure}{Production crop + forestry database}
    \end{center}
    
    \begin{center}
        \includegraphics[width=0.45\textwidth]{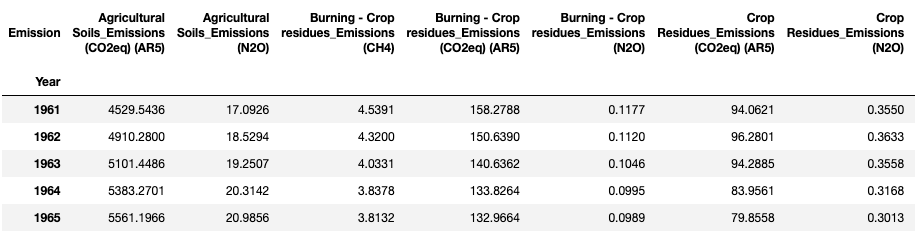}
        \captionof{figure}{Emissions database}
    \end{center}
    
    The preprocessed dataset that is fed into the VAR (Vector Auto Regressive) model is of the structure as shown in Figure 6.
    \end{enumerate}
    
    \begin{center}
        \includegraphics[width=0.50\textwidth]{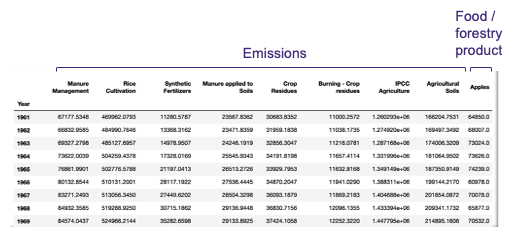}
        \captionof{figure}{Preprocessed data}
    \end{center}

    \subsection{Features (for every country and product)}
    
    \begin{enumerate}
        \item Year (from 1961 to 2019)
        \item Emission features (in kilotonnes) \\
            a. Manure management \\ 
            b. Rice cultivation \\
            c. Synthetic fertilizers \\
            d. Crop residues \\
            e. Up to total number of emission features that is based on country’s dataset
        \item One Production crop/Forestry product (in hg/ha)
    \end{enumerate}

    \subsection{Exploratory Data Analysis}
    EDA in this project is an approach used to analyze large data sets to understand how different features perform over time, often using statistical graphics and other data visualizations. EDA often helps us understand outliers or class imbalance, if any. Fig 7 shows the trend graphs for production of crop/livestock in Asia since 1961.
    
    \begin{center}
        \includegraphics[width=0.45\textwidth]{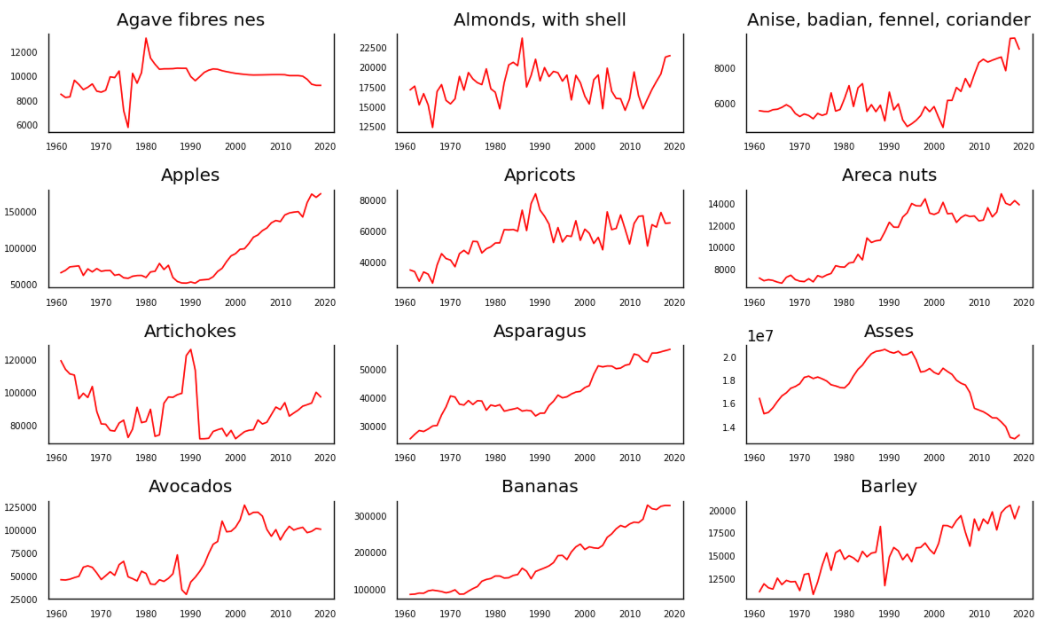}
        \captionof{figure}{Time series graphs for crop/livestock production}
    \end{center}

\section{Experiments}
Proposed model - Vector Auto regression model (VAR model)
VAR model is a multivariate forecasting regression algorithm used when two or more variables influence each other. Before a dataset is trained with the statistical model, the dataset needs to pass two tests which are as follows,

    \begin{enumerate}
        \item Granger Causality Test \\
        A statistics test to check whether one variable influences the prediction of another variable. It tests whether past values of x aid in the prediction of yt. If so, x is said to “Granger cause” y. Figure 8 shows the granger causation matrix of one production crop affected by emission gases. From the figure, if a p-value is less than the critical threshold of 0.05 or 0.01, the corresponding X series (column) causes the Y (row).
    
        \begin{center}
             \includegraphics[width=0.45\textwidth]{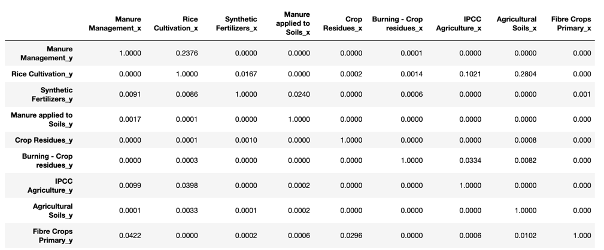}
            \captionof{figure}{Granger causation matrix for Fibre crops}
        \end{center}
        
        \item Augmented Dickey–Fuller (ADF) Test \\
        Statistical test used to test whether a given time series is stationary or not. A stationary time series has statistical properties or moments (e.g., mean and variance) that do not vary in time
        
        \begin{center}
             \includegraphics[width=0.45\textwidth]{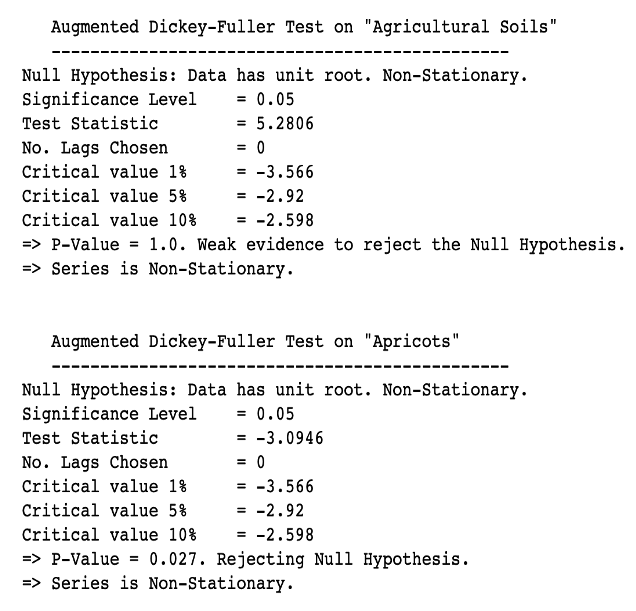}
             \captionof{figure}{ADF Test}
        \end{center}
        
        \begin{center}
             \includegraphics[width=0.35\textwidth]{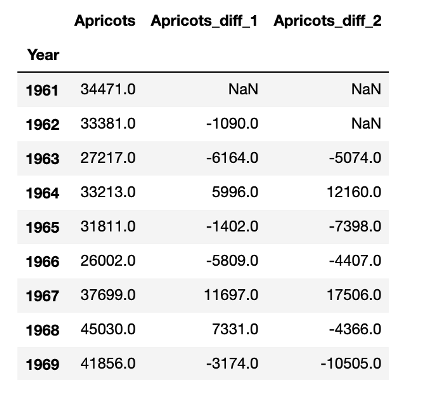}
            \captionof{figure}{ADF Test difference results}
        \end{center}
        
    \end{enumerate}
    
    Figure 9 shows the ‘apricots’ series that is non-stationary. Hence it is subject to first and second difference which results in a stationary series. \\
    VAR models are constructed based on their order, which refers to the number of historical time periods the model will use. For example, a 4th-order VAR would model each year's apricot production quantity as a linear combination of the last four years of apricot quantities. A pth-order VAR model is written as shown in equation 1
    \begin{equation}
    \ y_t = \gamma +A_1y_{t-1} +...+A_py_{t-p} +u_t ,t \in Z
    \end{equation}
    From the equation, yt indicates that variable's value i time periods earlier and is called the "ith lag" of yt. The variable gamma is a vector of constants which is the intercept of the model. Ap is a time-invariant (k × k)-matrix and ut is a k-vector of error terms.

    \subsection{Training and Inference}
    
        \begin{center}
                \includegraphics[width=0.25\textwidth]{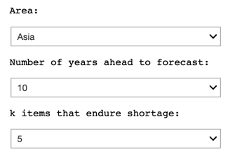}
                \captionof{figure}{User Inputs}
        \end{center}
    
        Both the training and inference of the data sets take place sequentially after the user input. The user can provide the area to forecast, the number of years ahead, and the top-k items that endure shortage in the specific year. The provided information pulls up the relevant dataset from the database. The datasets are cleaned, preprocessed, subject to two tests (ADF and Granger causality test), and fed as input to the VAR model with a lag order of 6. The model output is used to forecast the production quantities of all crops, livestock, or forestry products. The forecasted output values 50\% less than the mean of historical values are separated as they are more likely to have shortages. From this filtered set, the values are sorted by ‘abs(recentforecast – historicalmean),’ and the top-k items and their trend graphs are presented as outputs to the user. The outputs are shown in Figure 11 and Figure 12.
        
        \begin{center}
                \includegraphics[width=0.45\textwidth]{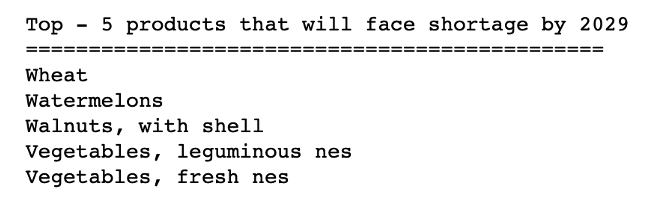}
                \captionof{figure}{Predicted outputs for Asia}
        \end{center}
        
        \begin{center}
                \includegraphics[width=0.45\textwidth]{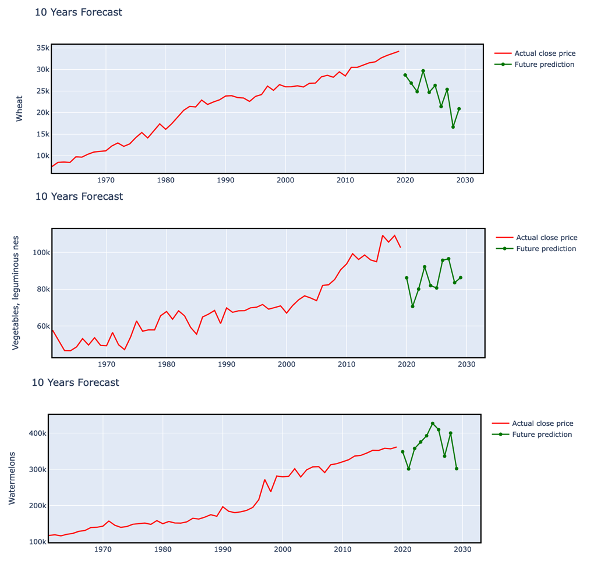}
                \captionof{figure}{10 year forecast for Asia}
        \end{center}

\section{Evaluation and Results}
In order to evaluate the performance of the trained model, the experiment is repeated differently. The dataset is split into a training set (50 historical values - 1961 to 2010) and an evaluation set (9 historical values - 2011 to 2019). The data is trained with a VAR model having a lag order of 6 using the 50 historical value dataset. The model is used to forecast the future nine values. The predicted values are compared against the evaluation set to calculate the evaluation metrics. Figure 13 shows the evaluation forecasting graphs for two production crops/livestock. From the trend, it is evident that the forecast is accurate to the actual production quantities.

    \begin{center}
            \includegraphics[width=0.35\textwidth]{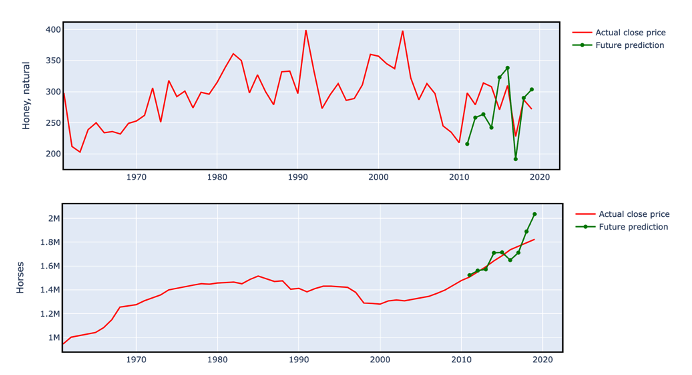}
            \captionof{figure}{Evaluation forecasting graphs}
    \end{center}
    
    The evaluation metrics for a time series regression are as follows,
    \begin{itemize}
        \item Mean absolute error (MAE) \\
        MAE is the average of the absolute difference between forecasted and true values. From equation 2, yi is the expected value, xi is the actual value and n represents the total number of values in the test set. The lower the MAE value, the more optimal the model.
    
        \begin{equation}
            \ MAE = (1 \div n) + \sum|y_i - x_i|
        \end{equation}
    
        \item Mean squared error (MSE) \\
        MSE is the average of the error squares.This metrics takes into account the variance (the difference between anticipated values) and bias (the distance of predicted value from its true value).
        
        \begin{equation}
            \ MSE = (1 \div n) + \sum{(y_i - x_i)}^2
        \end{equation}
        
        \item Root mean squared error \\
        This measure is the square root of mean square error. From equation 4, Where yi is the expected value, xi is the actual value and n represents the total number of values in the test set. All three error terms need to be as low as possible for the model to be called optimal.
        
        \begin{equation}
            \ RMSE = \sqrt(\sum{(y_i - x_i)}^2 \div n)
        \end{equation}
    
    The average performance metric for the trained VAR model with lag order of 6 is as follows,
    \begin{itemize}
        \item Mean absolute error - 0.359
        \item Mean squared error - 0.203
        \item Root mean squared error - 0.434
    \end{itemize}
    The experiment results show that the VAR model is a sufficiently good model to forecast the production of crops when it is under the influence of emission of gases.
    \end{itemize}

\section{Conclusion and Discussion}
The research project focuses on creating a system that can predict the production of crops, livestock and forestry products under the influence of gases. With over 30\% of global Greengrass gas emissions directly released due to food production, it helps humans realize the direct impact of food production and environmental pollution. The proposed solution uses the food and agriculture database from the United Nations to train a vector autoregressive model. 

The forecast model can project trends of production quantities and the top-k items that would endure shortage in the future. The results of the solution help in understanding the importance of sustainability to make the necessary lifestyle changes in developed countries before it is too late. In underdeveloped and developing countries, the results could help the government and agriculturists produce the right quantity of food products to satisfy the country's demand.

The future work of this project would focus on adding more datasets on Temperature change, Rainfall, Population, Trade flows that would add intricate factors to decide the production of
crops in the future. The solution can also be extended to cover social and economic sustainability to make the system a robust model for sustainability prediction. Technology companies like Microsoft and AWS are investing more in building sustainability solutions to reduce the carbon footprint. This research gives insights into what factors an organization should care about while designing sustainable solutions. Given the right dataset,this research can be further extended to be an in-built library to make sustainable factor predictions on social, environmental, and economic aspects.

\section{References}
\label{sec:reference_examples}
\nobibliography*
\bibentry{c:5}. \\\\
\bibentry{r:6}. \\\\
\bibentry{r:1}. \\\\
\bibentry{c:2}. \\\\
\bibentry{c:3}. \\\\
\bibentry{c:4}.

% Use \bibliography{yourbibfile} instead or the References section will not appear in your paper
\nobibliography{aaai23}

\end{document}